\documentclass[lettersize,journal]{IEEEtran}
\usepackage[perpage, hang, flushmargin]{footmisc}
\renewcommand{\footnoterule}{
    \kern -3pt
    \hrule width 2in height 0.4pt
    \kern 2.6pt
}
\usepackage{amsmath,amsfonts}
\usepackage{mathrsfs}
\usepackage{algorithmic}
\usepackage{flushend}
\usepackage{cite}
\usepackage{hyperref}
\usepackage{array}
\usepackage[caption=false,font=normalsize,labelfont=sf,textfont=sf]{subfig}
\usepackage{textcomp}
\usepackage{stfloats}
\usepackage{url}
\usepackage{verbatim}
\usepackage{graphicx}
\usepackage{amssymb}
\usepackage{booktabs}
\usepackage{tabu}
\usepackage{multirow}
\usepackage{bm}
\usepackage{indentfirst}
\usepackage[table,xcdraw]{xcolor}

\def\BibTeX{{\rm B\kern-.05em{\sc i\kern-.025em b}\kern-.08em
    T\kern-.1667em\lower.7ex\hbox{E}\kern-.125emX}}
\usepackage{balance}

\begin{document}

\title{Generalized Face Forgery Detection via Adaptive Learning for Pre-trained Vision Transformer}

\author{Anwei~Luo, Rizhao~Cai, Chenqi~Kong, Yakun~Ju, Xiangui~Kang~\IEEEmembership{Senior Member,~IEEE}, Jiwu~Huang~\IEEEmembership{Fellow,~IEEE} and~Alex~C.~Kot~\IEEEmembership{Life Fellow,~IEEE}%

\thanks{A. Luo is with the School of Information Technology, Jiangxi University of Finance and Economics, Nanchang 330013, China and was with the School of Computer Science and Engineering, Sun Yat-Sen University, Guangzhou 510006, China. (email: luoanw@mail2.sysu.edu.cn)}
\thanks{X. Kang is with the School of Computer Science and Engineering, Sun Yat-Sen University, Guangzhou 510006, China. (isskxg@mail.sysu.edu.cn)}
\thanks{R. Cai, C. Kong, Y. Ju and Alex C. Kot are with the Rapid-Rich Object Search (ROSE) Lab, School of Electrical and Electronic Engineering, Nanyang Technology University, Singapore. (e-mail: rzcai@ntu.edu.sg; chenqi.kong@ntu.edu.sg; kelvin.yakun.ju@gmail.com; eackot@ntu.edu.sg) }
\thanks{J. Huang is with the Guangdong Key Laboratory of Intelligent Information Processing and National Engineering Laboratory for Big Data System Computing Technology, Shenzhen University, Shenzhen 518060, China.}
\thanks{Alex C. Kot is also with the China-Singapore International Joint Research Institute, Singapore.}
\thanks{This work is partially done in the Rapid-Rich Object Search (ROSE) Lab, Singapore.}
}

\markboth{Journal of \LaTeX\ Class Files,~Vol.~18, No.~9, September~2020}%
{How to Use the IEEEtran \LaTeX \ Templates}

\maketitle

\begin{abstract}

With the rapid progress of generative models, the current challenge in face forgery detection is how to effectively detect realistic manipulated faces from different unseen domains. Though previous studies show that pre-trained Vision Transformer (ViT) based models can achieve some promising results after fully fine-tuning on the Deepfake dataset, their generalization performances are still unsatisfactory. One possible reason is that fully fine-tuned ViT-based models may disrupt the pre-trained features \cite{kumar2022finetuning, chen2022adaptformer} and overfit to some data-specific patterns \cite{ojha2023towards}. To alleviate this issue, we present a \textbf{F}orgery-aware \textbf{A}daptive \textbf{Vi}sion \textbf{T}ransformer (FA-ViT) under the adaptive learning paradigm, where the parameters in the pre-trained ViT are kept fixed while the designed adaptive modules are optimized to capture forgery features. Specifically, a global adaptive module is designed to model long-range interactions among input tokens, which takes advantage of self-attention mechanism to mine global forgery clues. To further explore essential local forgery clues, a local adaptive module is proposed to expose local inconsistencies by enhancing the local contextual association. In addition, we introduce a fine-grained adaptive learning module that emphasizes the common compact representation of genuine faces through relationship learning in fine-grained pairs, driving these proposed adaptive modules to be aware of fine-grained forgery-aware information. Extensive experiments demonstrate that our FA-ViT achieves state-of-the-arts results in the cross-dataset evaluation, and enhances the robustness against unseen perturbations. Particularly, FA-ViT achieves 93.83\% and 78.32\% AUC scores on Celeb-DF and DFDC datasets in the cross-dataset evaluation. The code and trained model have been released at: https://github.com/LoveSiameseCat/FAViT.




\end{abstract}


\section{Introduction}
\IEEEPARstart{W}{ith} the rapid development of deep learning technology, Artificial Intelligence-Generated Content (AIGC) technology has made significant progress in a wide variety of multimedia tasks. However, this advancement poses a grand challenge to the human eye in discriminating these digital contents. In particular, attackers can easily generate falsified facial content (a.k.a. \emph{Deepfakes}) for various malicious purposes, posing a pressing threat to society over financial fraud, political conflict, and impersonation. To protect the authentication integrity, it is of paramount importance to develop effective methods for face forgery detection.

Most previous works build their detectors using Convolutional Neural Network (CNN), of which Xception \cite{chollet2017xception} or EfficientNet \cite{tan2019efficientnet} is widely employed as the basic backbone due to their outstanding Deepfake detection performance. To enhance the generalizability, some works explore the common forgery clues hidden in manipulated faces, such as noise information \cite{luo2021generalizing, masi2020two, guo2023exposing}, blending artifacts \cite{li2020face, nguyen2024laa, shiohara2022detecting}, and frequency features \cite{xu2020learning, li2021frequency, qian2020thinking}, etc. However, the limited receptive field in CNNs constrains their capacity to comprehensively learn more generalized features \cite{liu2020global}. In response, some methods seek to use Vision Transformers (ViT) for face forgery detection \cite{xu2023tall, wang2022m2tr, liu2022multi, zhao2023istvt}. Due to the self-attention mechanism, these ViT-based methods can model the long-range relations among different input tokens. However, ViT struggles to capture the local feature details, which are particularly crucial in Deepfake detection \cite{miao2023f}. To address this limitation, previous works incorporate CNN local priors into ViT architectures \cite{zhuang2022uia, miao2022hierarchical}.

Pre-trained models are known to exhibit better convergence and generalization for downstream tasks as compared to models trained from scratch \cite{nakashima2022can}, and the pre-trained ViT \footnote[1]{Specifically refers to models trained on public large-scale vision datasets.} has been proven effective in forensic tasks\cite{ojha2023towards}. Therefore, in previous works \cite{miao2022hierarchical, miao2023f, zhuang2022uia, xu2023tall}, it is common to use publicly accessible pre-trained weights to initialize the ViT-based detectors, and subsequently update these parameters on Deepfake datasets. However, recent works \cite{kumar2022finetuning, chen2022adaptformer} point out that ViT-based models fully fine-tuned on specific down-stream tasks would disrupt pre-trained features and may overfit to data-specific patterns \cite{ojha2023towards}, potentially hindering their generalization capability in open-set environments. These observations suggest that developing a new model capable of capturing additional forgery-aware information without disrupting pre-trained knowledge is necessary for generalized face forgery detection.

\begin{figure*}[t]
    \centering
    \includegraphics[scale=0.67]{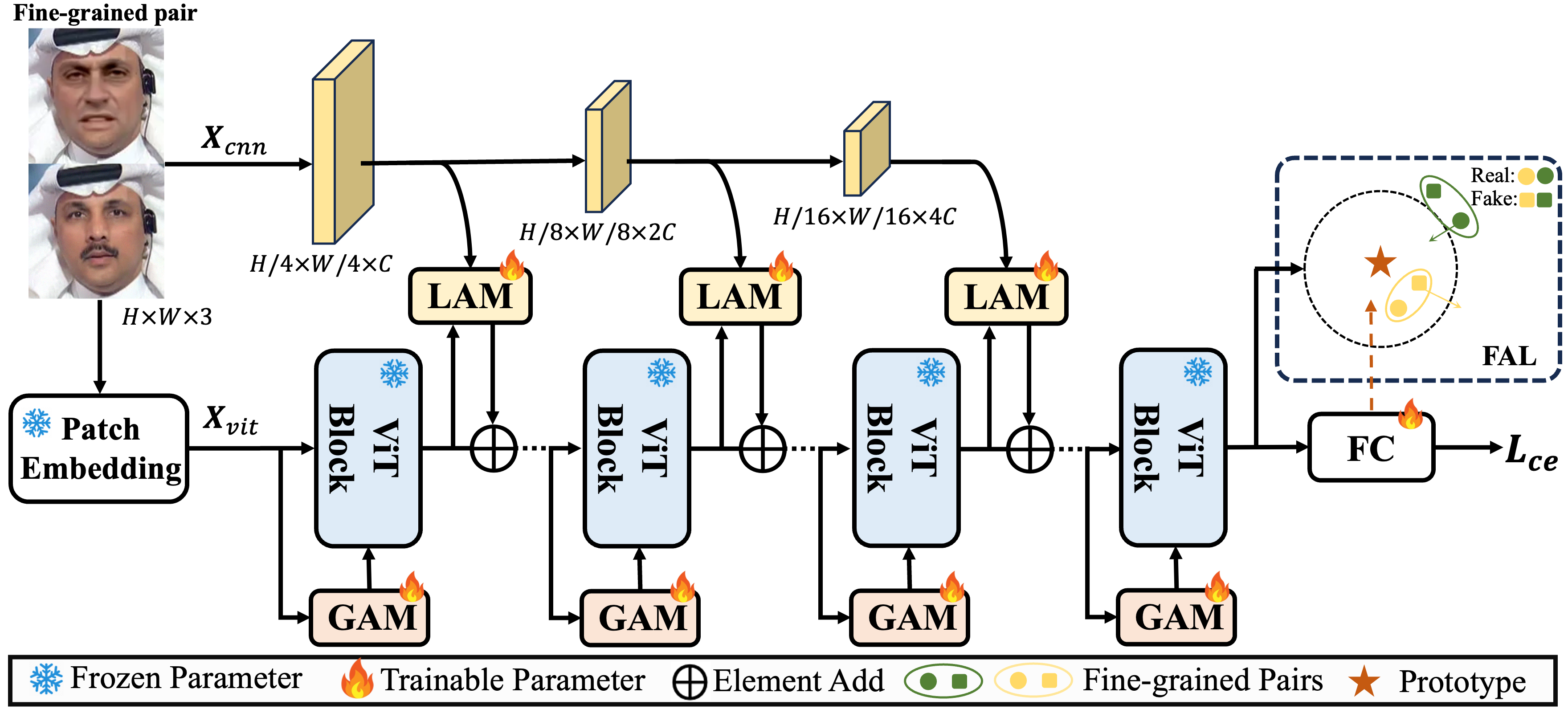}
   \caption{Overview of the proposed \textbf{F}orgery-aware \textbf{A}daptive \textbf{Vi}sion \textbf{T}ransformer (FA-ViT). Global Adaptive Modules (GAM) and Local Adaptive Modules (LAM) are inserted into the fixed pre-trained ViT to learn different types of forgery-aware information. Fine-grained Adaptive Learning (FAL) further guides these modules to mine fine-grained clues from subtle discrepancies in fine-grained pairs. We omit CNN blocks for better visualization.}
   \label{framework}
\end{figure*}

To this end, we propose the \textbf{F}orgery-aware \textbf{A}daptive \textbf{Vi}sion \textbf{T}ransformer (FA-ViT), where the pre-trained weights are fixed and only the designed adaptive modules are optimized to capture abundant forgery-aware information. The overview of FA-ViT is illustrated in Fig.~\ref{framework}. The Global Adaptive Module (GAM) is integrated into the self-attention layer, and interacts with original query, key, and value tokens. Leveraging the self-attention mechanism, GAM focuses on learning forgery-aware information in a global view. On the other hand, the Local Adaptive Module (LAM) employs quadratic encoding \cite{Cordonnier2020On} to enhance the local context between each ViT token and its neighboring spatial information. It complementarily explores the local information to improve ViT's expressivity. As such, the global and local forgery-aware information jointly adapt to the fixed pre-trained features, forming the generalized forensic representation for detecting Deepfakes across various scenarios.

Furthermore, the commonly used cross-entropy loss emphasizes category-level differences but struggles to capture the fine-grained information that reveals subtle differences between manipulated and genuine faces \cite{sun2022dual, lu2023detection, li2021frequency}. Therefore, we design the Fine-grained Adaptive Learning (FAL) to guide the proposed adaptive modules in mining these critical information. As shown in Fig.~\ref{framework}, the fine-grained pairs, which share similar visual semantic content but belong to different categories, are grouped together as input pairs. FAL utilizes the weight from the last Fully Connected (FC) layer as the proxy prototype of genuine faces, and regularizes the relationship between the prototype and each fine-grained pair through circle loss \cite{sun2020circle}, thereby exposing more fine-grained forgery-aware information during the adaptive learning process.

Our main contributions are summarized as follows:

\begin{itemize}

\item We observe that ViT-based models struggle to generalize to unseen datasets when fully fine-tuning on the task of Deepfake detection. To address this issue, we propose the novel Forgery-Aware Adaptive Vision Transformer (FA-ViT) tailored for generalized face forgery detection under the adaptive learning paradigm.

\item We propose the Global Adaptive Module (GAM) and the Local Adaptive Module (LAM), which effectively adapt global and local forgery-aware information to the pre-trained ViT features. In addition, we design a novel Fine-grained Adaptive Learning (FAL) to guide these adaptive models to capture fine-grained information in the adaptive learning process.

\item We conduct extensive experiments on multiple datasets and senarios, and the results demonstrate our proposed FA-ViT outperforms state-of-the-art methods in a wide variety of evaluations.

\end{itemize}

\section{Related Work}
\subsection{Face Forgery Detection}

Early attempts in face forgery detection mainly relied on hand-crafted features, such as visual artifacts \cite{matern2019exploiting}, face wrapping distortions \cite{li2018exposing}, and abnormal blinking frequency \cite{li2018ictu}. However, these methods suffer from limited detection accuracy. Recent studies design neural networks for face forgery detection, like MesoNet \cite{afchar2018mesonet}, CapsuleNet \cite{nguyen2019capsule}, Xception \cite{chollet2017xception}, and EfficientNet \cite{tan2019efficientnet}. However, these CNN-based methods are prone to overfitting to the training data, making them difficult to generalize well to unseen datasets. To improve the generalization performance, follow-up works \cite{yin2024improving, yu2023narrowing, yu2023augmented} delve into domain expert knowledge to guide the model in learning generalized representations. Face X-ray \cite{li2020face}, SBI \cite{shiohara2022detecting}, and LAA-Net \cite{nguyen2024laa} build well-annotated faces in an unsupervised manner, but the learned blending artifacts can be easily eliminated by common perturbations. SCL \cite{li2021frequency}, RECCE \cite{cao2022end}, DCL \cite{sun2022dual}, Lisiam \cite{wang2022lisiam}, and CFM \cite{luo2023beyond} explore auxiliary information to learn more critical forgery features, while F3Net \cite{qian2020thinking} and SPSL \cite{liu2021spatial} reveal manipulation traces beyond RGB space. LSDA\cite{yan2024transcending} diversifies forgery types by augmenting forgery features in the latent space. Although these local features extracted from CNN are effective in face forgery detection, the limited receptive field constrains their capability to learn global representations \cite{naseer2021intriguing, peng2021conformer}.

In contrast, ViT-like architectures model the global relations through self-attention mechanism and capture long-distance information in the final representation. Due to the lack of inductive bias, ViT often ignore the local feature details, which are rather critical in face forgery detection \cite{zhao2021multi, gu2022exploiting, miao2023f}. This issue can be mitigated by introducing local prior knowledge into ViT. For example, UIA \cite{zhuang2022uia} supervises the attention map generated from self-attention layers through local consistency learning. 
M2TR \cite{wang2022m2tr}, HFI-Net \cite{miao2022hierarchical}, and F2trans \cite{miao2023f} combine CNN and ViT features to learn complementary information in the final representation. However, these ViT-based methods are designed in the fully fine-tuning paradigm, which may lead to the loss of valuable pre-trained knowledge. To solve this, our proposed FA-ViT utilizes additional modules to adaptively learn forgery-aware information while keeping ViT parameters preserved, resulting in a better representation learning and enhanced detection performance.

\subsection{Model Adaptation}
In general, fine-tuning all model parameters is widely used for downstream tasks. However, recent works \cite{kumar2022finetuning, chen2022adaptformer} have demonstrated that the fully fine-tuning strategy may disrupt pre-trained features and does not generalize well to out-of-distribution data. As pre-trained models become larger and stronger, the efficiency and stability issues of the fully fine-tuning strategy become more severe. Ojha \emph{et. al.} \cite{ojha2023towards} reveal that training only a learnable classifier for a fixed CLIP-ViT achieves better generalization performance than fully fine-tuning strategy in synthetic image detection. However, this is a sub-optimal solution since the fixed model lacks task-specific knowledge. To explore new model adaptation strategies, Lee \emph{et. al.} \cite{lee2023surgical} propose an alternative strategy by selectively fine-tuning a subset of layers, which demonstrated a better performance for image corruption task. Nevertheless, manually selecting the optimal layers for fine-tuning is challenging and time-consuming. To address this issue, Side-Tuning \cite{zhang2020side}, to the best of our knowledge, is the first work to explore model adaptation by using additive modules with fixed pre-trained models. It established a baseline and verified the effectiveness across different tasks. Subsequently, this adaptation strategy has been successfully applied to different tasks with specific expert design, including video action recognition \cite{pan2022st}, natural language understanding \cite{hu2022lora}, semantic segmentation \cite{xu2023side}, and anti-spoofing \cite{huang2022adaptive}. Our experiments in Sec.~\ref{section_model_ablation} also demonstrated that fully fine-tuning the pre-trained ViT model leads to sub-optimal performance compared to other model adaptation strategies. Grounded in these observations, we customize a novel adaptive learning scheme on the pre-trained ViT backbone to achieve more generalized face forgery detection.

\subsection{Fine-grained Information Learning}

As counterfeits become increasingly realistic, formulating face forgery detection as a fine-grained classification task \cite{zhao2021multi, lu2023detection, gu2022exploiting} is preferable to expose local and subtle forgery traces. Zhao \emph{et al.} \cite{zhao2021multi} propose a multi-attentional learning framework that extracts local discriminative features from different attention maps. Lu \emph{et al.} \cite{lu2023detection} further extend this attention mechanism into time domain to expose long distance inconsistencies. SFDG \cite{wang2023dynamic} explores content-adaptive features by fusing fine-grained and coarse-grained frequency information via dynamic graph learning. MTD-Net \cite{yang2021mtd} and F2Trans \cite{miao2023f} utilize the central difference operation \cite{yu2020searching} to reveal fine-grained forgery clues. The aforementioned works have demonstrated that fine-grained information is crucial for achieving generalized face forgery detection. In our proposed adaptive learning paradigm, FAL is proposed to drive these adaptive modules to focus on subtle discrepancies in each fine-grained pair, thereby facilitating the learning of fine-grained information from the perspective of the optimization objective.

\section{Proposed Method}

In this section, we first provide an overview of the proposed FA-ViT in Sec.~\ref{overview}. Then we describe the Global Adaptive Module (GAM) and Local Adaptive Module (LAM) in Secs.~\ref{intro_GAM} and ~\ref{intro_LAM}, respectively. Finally, the Fine-grained Adaptive Learning (FAL) is presented in Sec.~\ref{intro_FAL}.

\subsection{Overview} \label{overview}
The framework of our proposed FA-ViT is illustrated in Fig.~\ref{framework}. It adopts the pre-trained ViT as the basic backbone, which comprises twelve ViT blocks. Each block consists of a Self-Attention (SA) layer and a Multi-Layer Perceptron (MLP) layer. During training, the parameters within these blocks remain frozen.

In FA-ViT, the input $\mathbf{X}\in\mathbb{R} ^{H\times W\times 3}$ is divided into $L$ image patches and further processed into $D$ dimensional tokens, denoted as $\mathbf{X}_{vit}\in\mathbb{R} ^{L\times D}$. In each ViT block, we insert a GAM into the SA layer. When $\mathbf{X}_{vit}$ passes through the SA layer, GAM captures the global forgery-aware information with the help of self-attention mechanism. On the other hand, we extract multi-scale spatial features from $\mathbf{X}$, where each scale feature $\mathbf{X}_{cnn}$ is obtained from a CNN block consisting of three convolutional layers. LAM aggregates spatial forgery-aware information from $\mathbf{X}_{cnn}$ for each ViT token, which benefits in capturing rich local details in an adaptive learning manner. In addition to the commonly used Cross Entropy (CE) loss, we introduce FAL to guide GAMs and LAMs to capture more fine-grained forgery-aware information.

\begin{figure}[t]
  \centering
   \includegraphics[width=3.6 in]{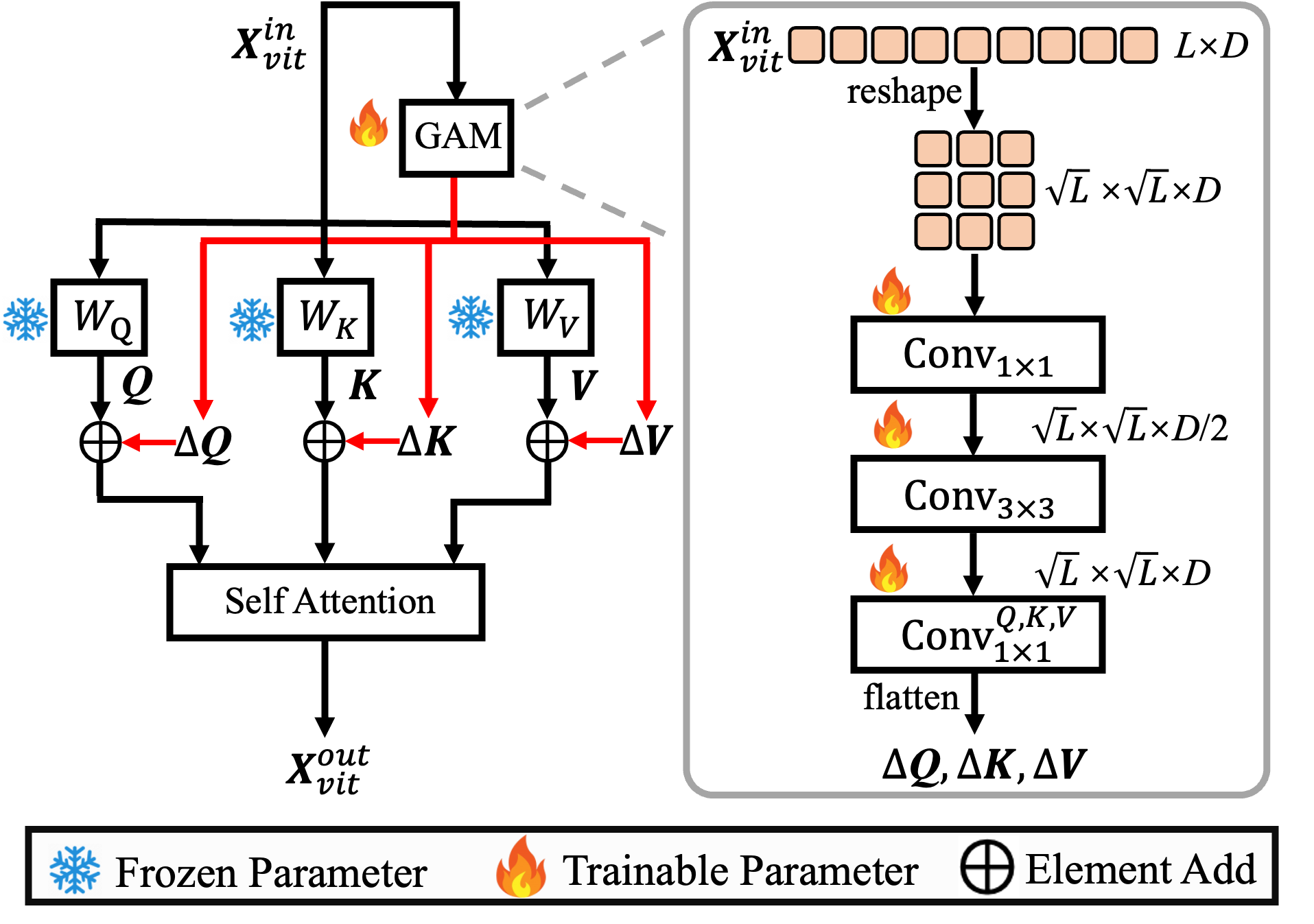}

   \caption{The details of the proposed global adaptive module.}
   \label{GAM}
\end{figure}

\subsection{Global Adaptive Module} \label{intro_GAM}

The self-attention layer is a key component in ViT, which enables each input token to aggregate information from all other tokens. Our proposed GAM is built on this layer to mine global information by using the self-attention mechanism. We first briefly introduce the calculation process in self-attention layer as follow.

 Denote $\mathbf{X}^{in}_{vit}\in\mathbb{R} ^{L\times D}$ is the input of the ViT block, it is first projected to query tokens $\mathbf{Q} \in\mathbb{R} ^{L\times D}$, key tokens $\mathbf{K} \in\mathbb{R} ^{L\times D}$ and value tokens $\mathbf{V} \in\mathbb{R} ^{L\times D}$ through three learnable matrices ${W}_{Q} \in\mathbb{R} ^{D\times D}$, ${W}_{K} \in\mathbb{R} ^{D\times D}$ and ${W}_{V} \in\mathbb{R} ^{D\times D}$:
\begin{equation}
\mathbf{Q} = \mathbf{X}^{in}_{vit}{W}_{Q}, ~\mathbf{K} = \mathbf{X}^{in}_{vit}{W}_{K}, ~\mathbf{V} = \mathbf{X}^{in}_{vit}{W}_{V}.
\end{equation}
Then the calculation in self-attention layer is expressed as:
\begin{equation}
    \mathbf{X}^{out}_{vit} = {\rm Attention}(\mathbf{Q},\mathbf{K},\mathbf{V})={\rm softmax}({\mathbf{Q}\mathbf{K}^\top}/{\sqrt{D}})\mathbf{V},
\end{equation}
where $\mathbf{X}^{out}_{vit}\in\mathbb{R} ^{L\times D}$ is the output.

The detail of our proposed GAM is shown in Fig.~\ref{GAM}. It is a bottleneck structure consisting of three convolutional layers, and flexibly inserted into the self-attention layer in a parallel manner. Specifically, $\mathbf{X}^{in}_{vit}$ is first reshaped into the shape of $\sqrt{L} \times \sqrt{L} \times D$ according to its original spatial position. Then a $1 \times 1$ convolutional layer is applied to reduce the dimension, followed by a $3 \times 3$ convolutional layer to capture the token-level dependency. Finally, GAM generates global adaptive information for $\mathbf{Q}$, $\mathbf{K}$, and $\mathbf{V}$ through three different $1 \times 1$ convolutions. This process is formulated as follows:
\begin{equation}
\bigtriangleup  {\mathbf{Q}},\bigtriangleup  {\mathbf{K}},\bigtriangleup  {\mathbf{V}} = \mathrm{Conv}_{1 \times 1}^{Q, K, V}(\mathrm{Conv}_{3 \times 3}(\mathrm{Conv}_{1 \times 1}( \mathbf{X}^{in}_{vit}))),
\end{equation}
\begin{equation}
     \mathbf{X}^{out}_{vit} = {\rm Attention}(\mathbf{Q} + \bigtriangleup{\mathbf{Q}}, \mathbf{K} + \bigtriangleup{\mathbf{K}}, {\mathbf{V}} + \bigtriangleup{\mathbf{V}}),
\end{equation}
where $\bigtriangleup{\mathbf{Q}}$, $\bigtriangleup{\mathbf{K}}$, and $\bigtriangleup{\mathbf{V}}$ are the adaptive information for original $\mathbf{Q}$, $\mathbf{K}$, and $\mathbf{V}$. Since the original and the adaptive information are fused together and subsequently processed by the self-attention operation, this ensures that GAM interacts with all the tokens to capture forgery-aware features from a global perspective.

\subsection{Local Adaptive Module} \label{intro_LAM}

\begin{figure}[t]
  \centering
   \includegraphics[width=3.4 in]{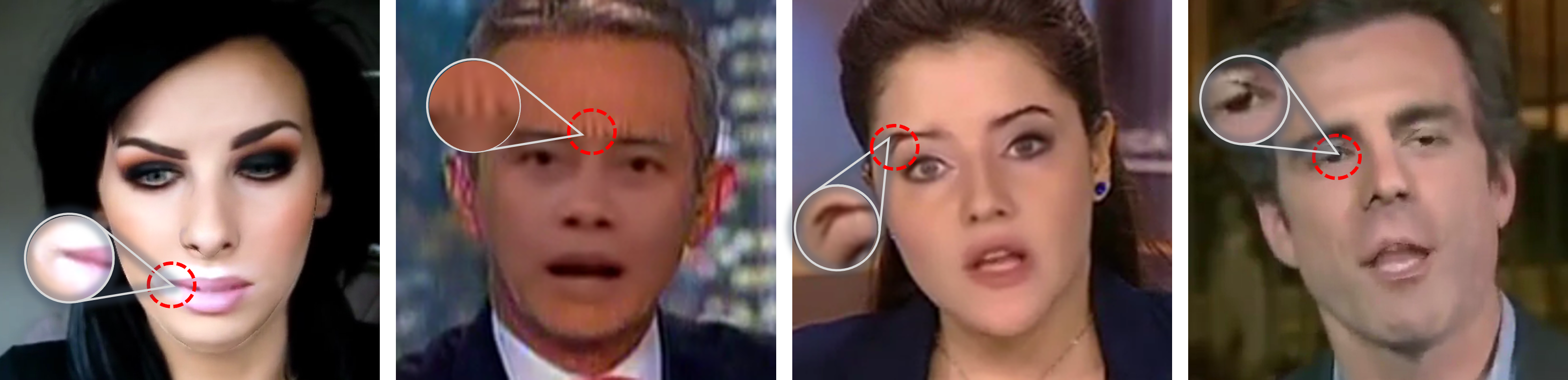}
   \caption{Examples of artifacts in manipulated faces that cause local inconsistencies.}
   \label{artifact}
   \vspace{-5 mm}
\end{figure}

The pure ViT processes input through stacked linear layers, which struggles to capture local details crucial for detecting manipulated faces. As shown in Fig.~\ref{artifact}, artifacts in manipulated faces often introduce local inconsistencies, indicating that each query token must emphasize its relationships with surrounding positions to detect local forgery clues. Previous approaches use cross-attention modules \cite{song2022face, liu2022multi, wang2022m2tr} or addition operations \cite{guan2022delving, miao2023f} to inject local spatial information into ViT-like architectures but often overlook the contextual importance of each query token. To alleviate this issue, our proposed LAM is designed to emphasize contextual information in the adaptive learning process, thus efficiently capturing critical local forgery clues from local inconsistencies.

\begin{figure}[t]
  \centering
   \includegraphics[width=3.4 in]{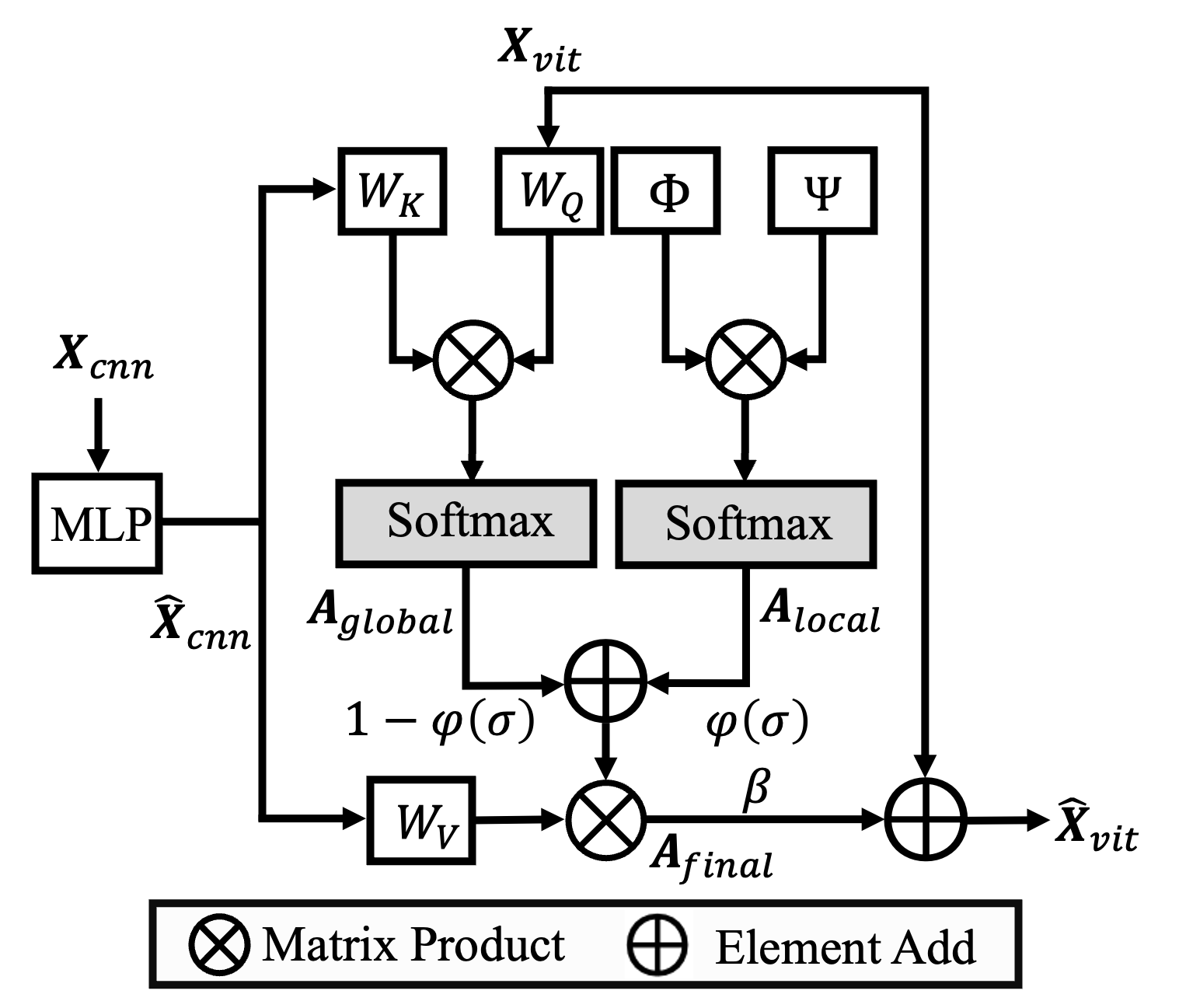}

   \caption{The details of the proposed local adaptive module.}
   \label{LAM}
\end{figure}

Fig.~\ref{LAM} provides the details of our proposed LAM. The spatial feature $\mathbf{X}_{cnn}$ is first projected to $\hat{\mathbf{X}}_{cnn}$ through a MLP layer, where $\hat{\mathbf{X}}_{cnn}$ and $\mathbf{X}_{vit}$ share the same shape. For the $a$th token $\mathbf{X}_{token}^{a}$ in $\mathbf{X}_{vit}$, LAM calculates its attention score at different parts of $\hat{\mathbf{X}}_{cnn}$ by simultaneously considering its global attention $\mathbf{A}_{global}^{a}$ and local attention $\mathbf{A}_{local}^{a}$:
\begin{equation}
\mathbf{A}_{final}^{a} = (1-\varphi(\sigma))\mathbf{A}_{global}^{a} + \varphi(\sigma)\mathbf{A}_{local}^{a},
\end{equation}
where $\mathbf{A}_{final}^{a}$ is the final attention map for $\mathbf{X}_{token}^{a}$. $\varphi(\cdot)$ represents sigmoid function, and $\sigma$ is a learnable parameter with zero initialization. The calculation of $\mathbf{A}_{global}^{a}$ is similar to cross-attention, which is expressed as follows: 
\begin{equation}
\mathbf{A}_{global}^{a} = softmax((\mathbf{X}_{token}^{a}W_{Q})(\hat{\mathbf{X}}_{cnn}W_{K})^{T}).
\end{equation}

\begin{figure}[t]
  \centering
   \includegraphics[width=3.4 in]{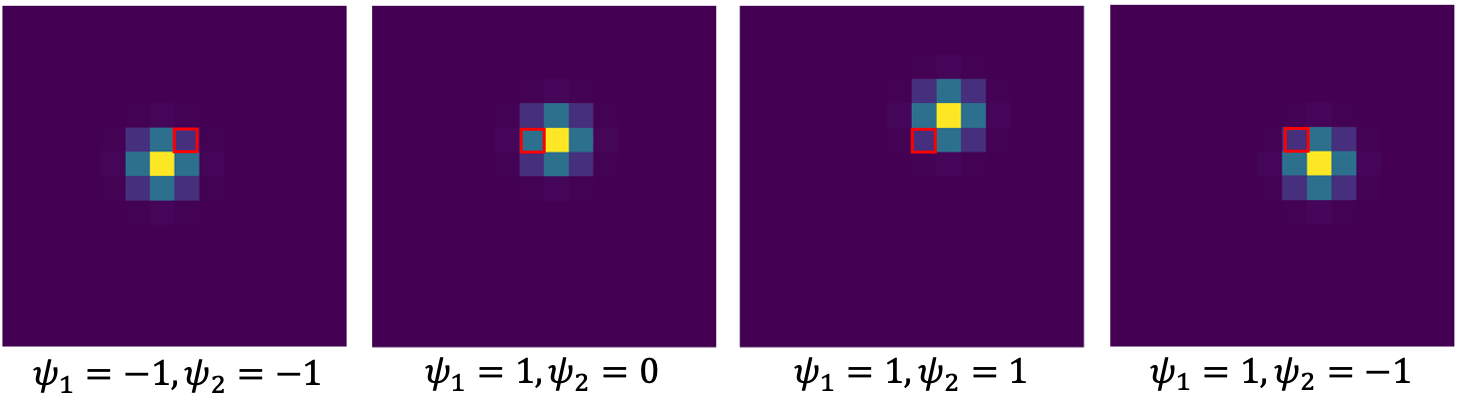}

   \caption{The local attention map of the query token (marked by a red box) under different combination of $\psi_{1}$ and $\psi_{2}$.}
   \label{Local_att}
\end{figure}

$\mathbf{A}_{local}^{a}$ is introduced to emphasize the locality of surrounding spatial information for $\mathbf{X}_{token}^{a}$ by using quadratic encoding \cite{Cordonnier2020On}:
\begin{equation}
\mathbf{A}_{local}^{a} = softmax(\Phi \Psi^{T}),
\end{equation}
where $\Phi \in \mathbb{R} ^{\sqrt{L}\times \sqrt{L}\times 3}$ represents the locality strength, and $\Psi \in \mathbb{R}^{1 \times 3}$ is the directional vector which determines the attention direction of $\mathbf{X}_{token}^{a}$. Note that we omit flatten operation before softmax operation for simplify. Supposing the spatial position of $\mathbf{X}_{token}^{a}$ is $(i_{a},j_{a})$, and similarly, the spatial position of $b$th token in $\hat{\mathbf{X}}_{cnn}$ is $(i_{b},j_{b})$. The relative locality strength $\phi_{a,b} \in \mathbb{R}^{1 \times 3}$ in position $(a,b)$ of $\Phi$ is expressed as:
\begin{equation}
\phi_{a,b} = \left (\left \| (i_{b}-i_{a}, j_{b}-j_{a}) \right \|_{2}, i_{b}-i_{a}, j_{b}-j_{a} \right).
\end{equation}
On the other hand, the directional vector $\Psi$ is expressed as:
\begin{equation}
\Psi=(-1,2\psi_{1},2\psi_{2}),    (\psi_{1},\psi_{2}) \in \left \{-1, 0, 1  \right \}. 
\end{equation}
In practice, we assign different values to $\psi_{1}$ and $\psi_{2}$ at different heads to explore local information in various directions, as illustrated in Fig.~\ref{Local_att}. We gather the final attention map for each query token to form $\mathbf{A}_{final}$ for $\mathbf{X}_{vit}$, and use it to inject local spatial information into ViT features. This process is formulated as follows:
\begin{equation}
\hat{\mathbf{X}}_{vit} = \mathbf{X}_{vit} + \beta \mathbf{A}_{final} \hat{\mathbf{X}}_{cnn}W_{V},
\end{equation}
where $\beta$ is a learnable scaling factor with zero initialization, and $\hat{\mathbf{X}}_{vit}$ is passed to next ViT block. Empirically, the multi-scale spatial information are injected into the first, fourth and seventh ViT block, respectively.

\subsection{Fine-grained Adaptive Learning} \label{intro_FAL}


Fine-grained information is very important for improving the generalization performance \cite{zhao2021multi, gu2022exploiting, miao2023f}. Thus we introduce FAL to facilitate the adaptive learning of fine-grained forgery-aware information. We first set the weight vector corresponding to the genuine face from the last FC layer as the proxy prototype of genuine faces, as previous work \cite{liu2018transductive} has demonstrated that the weights in classifier converge to the central direction of each class. In each fine-grained pair, FAL pulls closer the similarity between the prototype and the genuine face while pushing away the manipulated face from the prototype through the modified circle loss \cite{sun2020circle}:
\begin{equation}
L_{FAL}= \log{[ 1+  \sum \exp(\eta(\gamma_{n}(s_{n}-m_{n}) -\gamma_{p}(s_{p}-m_{p})))] }, 
\end{equation}
\begin{equation}
s_{p} = \mathrm{CosSim}(\mathbf{F}_{real}, \mathbf{F}_{pro}), ~s_{n} = \mathrm{CosSim}(\mathbf{F}_{fake}, \mathbf{F}_{pro})
\end{equation}
\begin{equation}
\gamma_{p} = max(1+m-s_{p},0),~\gamma_{n} = max(m+s_{n},0),
\label{FAL:eq1}
\end{equation}
\begin{equation}
m_{p} = 1-m,~m_{n} = m,
\label{FAL:eq2}
\end{equation}
where $\mathrm{CosSim}$ represents cosine similarity, and $\mathbf{F}_{pro}$ is the prototype of genuine faces. $\eta$ is the scaling factor. $m$ is a hyper-parameter to control the margin and weighting factors in fine-grained pairs. $\mathbf{F}_{real}$ and $\mathbf{F}_{fake}$ are encoded features of fine-grained pairs from the last ViT block.

\begin{figure}[!t]
    \centering
    \includegraphics[width=3.5 in]{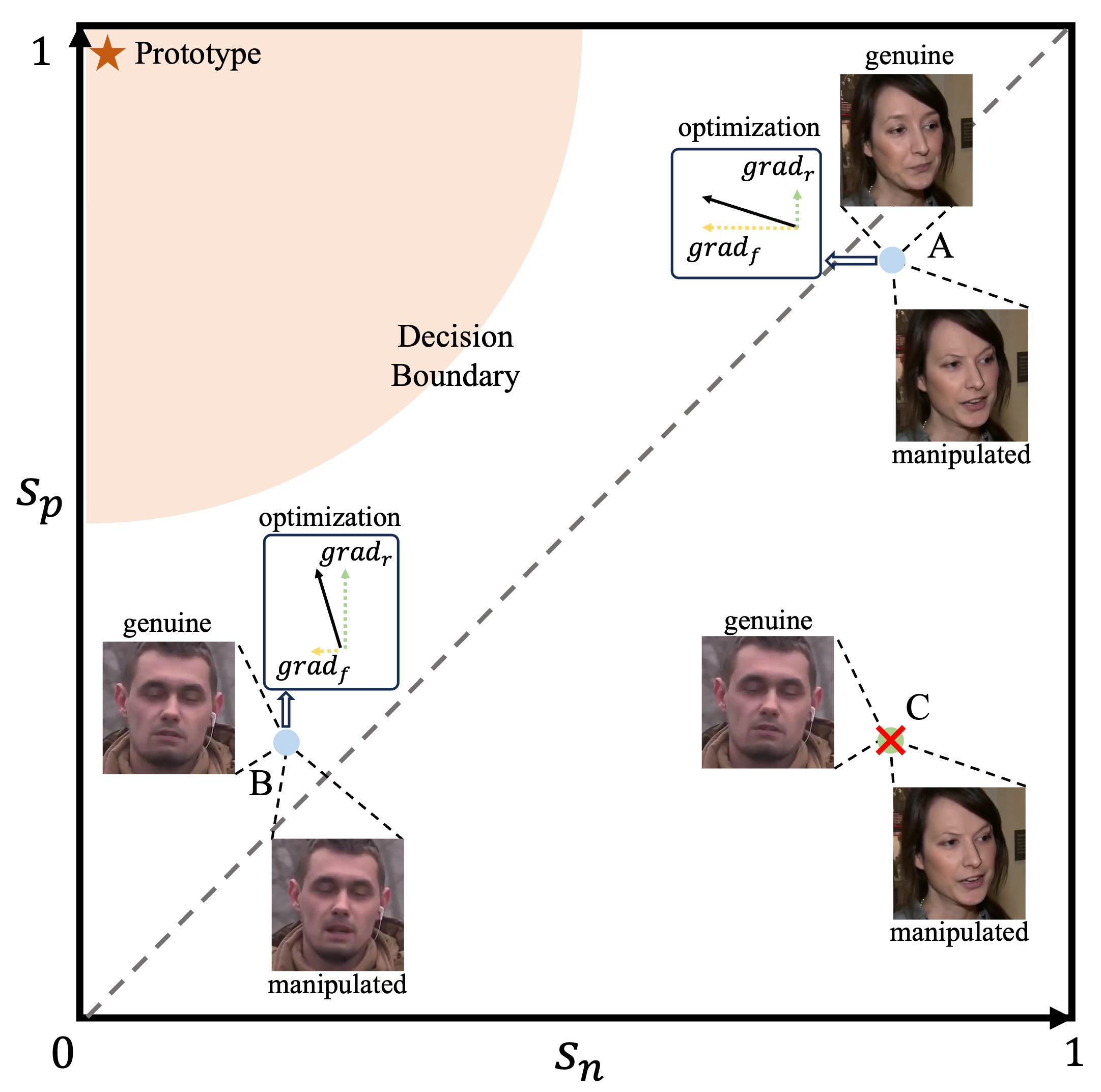}
   \caption{Optimization in fine-grained adaptive learning. The weight vector corresponding to the genuine face in the FC layer is used as the prototype of genuine faces. $grad_{r}$ and $grad_{f}$ are the gradients from genuine (real) and manipulated (fake) faces, respectively. (Best viewed in color)}
   \label{loss}
\end{figure}

In the optimization, the decision boundary is achieved at $\gamma_{n}( s_{n}-m_{n}) -\gamma_{p}(s_{p}-m_{p}) = 0$. By using Eqns.~\ref{FAL:eq1} and \ref{FAL:eq2}, the decision boundary is transformed to:
\begin{equation}
s_{n}^{2} + (1-s_{p})^{2} = 2m^{2}
\label{FAL:eq3}
\end{equation}
Eqn.~\ref{FAL:eq3} indicates that FAL encourages fine-grained pairs to converge towards a circle centered at $s_{p}=1$ and $s_{n}=0$ with a radius of $\sqrt{2}m$, as illustrated in Fig.~\ref{loss}. This optimization enables each fine-grained pair to provide a more flexible gradient to guide the learning of fine-grained information. For example, at point A, when $grad_{f}$ is much larger than $grad_{r}$, the model emphasizes capturing fine-grained discriminative information to push away the manipulated face. In contrast, when $grad_{f}$ is much smaller than $grad_{r}$ at point B, the model focuses on learning fine-grained consistency to pull the genuine face closer. On the other hand, FAL ignores the gradient from non-fine-grained pairs, such as the example of point C in Fig.~\ref{loss}. In this case, the discrepancy involves background or non-essential manipulation regions, which could potentially lead the model to overfit to trivial features. 

Obviously, FAL compact genuine faces in the feature space by exploring fine-grained information in subtle discrepancy regions. It does not explicitly penalize the distance of manipulated faces, as we want to preserve the diversity of manipulation traces in different forgery technologies. Unlike Single Center Loss (SCL) \cite{li2021frequency}, which uses the mean distance between different categories to compact the intra-class variance, FAL focuses on learning both fine-grained discriminative and consistent information in each fine-grained pair, making the process of mining critical information more precise.







\subsection{Total Loss}

our proposed FA-ViT is trained end-to-end and supervised by the cross-entropy loss between the prediction result {$\hat{y}$} and the ground truth label {$y$}:
\begin{equation}
    L_{ce}= -y\log_{}{\hat{y}}-(1-y)\log_{}{(1-\hat{y})},
\end{equation} 
where the label $y$ is 0 for genuine faces, otherwise $y$ is 1. The overall objective function consists of two components:
\begin{equation}
  L_{total}= L_{ce} + \lambda L_{FAL} , 
\end{equation}
where $\lambda$ is a weighting parameter. In the first training epoch, we set $\lambda$ to 0, allowing the model to focus on learning classification information. Subsequently, we adjust $\lambda$ to 1, thereby facilitating the learning of fine-grained information.

\section{Experiments}
\subsection{Experimental Setups}
\subsubsection{Dataset}

\begin{table}[]
\caption{The details of datasets used for training and testing.}
\label{dataset}
\centering
\begin{tabular}{ccc}
\toprule
Dataset                   & Usage & Real/Fake Videos \\ \hline
FaceForensics++ (FF++) \cite{rossler2019faceforensics++}    & Train & 720/2880        \\
DeeperForensics-1.0 (DFR) \cite{jiang2020deeperforensics}    & Train & 799/799        \\
FaceForensics++ (FF++) \cite{rossler2019faceforensics++}    & Test  & 140/560          \\
Celeb-DF-v2 (CDF) \cite{li2020celeb}        & Test  & 178/340          \\
WildDeepfake (WDF) \cite{zi2020wilddeepfake}        & Test  & 396/410          \\
DFDC-P \cite{ dolhansky2019deepfake}                     & Test  & 276/504          \\
DFDC \cite{ dolhansky2020deepfake}                     & Test  & 2500/2500          \\
DeepFakeDetection (DFD) \cite{dfd.org}    & Test  & 363/3431         \\
DeeperForensics-1.0 (DFR) \cite{jiang2020deeperforensics} & Test  & 201/201          \\ 
FFIW \cite{zhou2021face} & Test  & 2500/2500          \\ \bottomrule
\end{tabular}
\end{table}

We adopt eight widely used public datasets to evaluate our model. 1) \textbf{FaceForensics++} (FF++) \cite{rossler2019faceforensics++} is a widely used dataset which consists of four types of face manipulation techniques: DeepFakes (DF) \cite{hm16_20}, Face2Face (F2F) \cite{dfcode}, FaceSwap (FS) \cite{thies2016face2face}, and NeuralTextures (NT) \cite{thies2019deferred}. 2) \textbf{Celeb-DF-v2} (CDF) \cite{li2020celeb} is a high-quality Deepfake dataset specifically centered on celebrity faces. 3) \textbf{WildDeepfake} (WDF) \cite{zi2020wilddeepfake} collects the Deepfake videos from the internet, which includes diverse scenes and forgery methods. The evaluation results on WDF reflect the detector's performance in real-world scenarios. 4) \textbf{Deepfake Detection Challenge} (DFDC) \cite{dolhansky2020deepfake} provides a challenging dataset with a variety of Deepfake videos stemming from different scenarios, employing diverse Deepfake, GAN-based, and traditional face-swapping methods. 5) \textbf{Deepfake Detection Challenge Preview} (DFDC-P) \cite{dolhansky2019deepfake} provides a preview version of DFDC dataset and incorporates two facial modification algorithms. 6) \textbf{DeepFakeDetection} (DFD) \cite{dfd.org} serves as another comprehensive Deepfake dataset. This dataset comprises over 3,000 manipulated videos involving 28 actors and varying scenes. 7) \textbf{DeeperForenics-1.0} (DFR) \cite{jiang2020deeperforensics} is created by using real videos from FF++ with an innovative end-to-end face swapping framework. It also serves as a popular dataset to measure the morel's robustness. 8) \textbf{FFIW-10K} (FFIW) \cite{zhou2021face} is a recent large-scale dataset which focuses on multi-person scenarios. For the DFD dataset, we use all the videos for evaluation. For the other datasets, we follow the official strategy to split the corresponding datasets. For a comprehensive overview, please refer to Table \ref{dataset}.

\subsubsection{Implementation Details}
We use MTCNN \cite{zhang2016joint} to crop face regions and resize them to $224 \times 224$. We sample only 20 frames from each video to build the training data. Our method is implemented with PyTorch library \cite{paszke2019pytorch} on a single NVIDIA GTX 3090. We adopt the ViT-Base model pre-trained on ImageNet-21K \cite{ridnik2021imagenet} as the main backbone in FA-ViT. For optimization, we employ Adam \cite{kingma2014adam} optimizer with an initial learning rate of $3\times10^{-5}$, a weight decay of $1\times10^{-5}$, and a batch size of 32. The learning rate decays by 0.5 every 5 epochs. The hyper-parameters of $m$ and $\eta$ in FAL will be discussed in Sec. ~\ref{section_ablation}.

\subsubsection{Evaluation Metrics}
We follow the evaluation strategy in \cite{miao2023f}, the Accuracy (ACC) and Area Under the receiver operating characteristic Curve (AUC) are used as our evaluation metrics. For a fair comparison, we average the predictions in the same video to obtain the video-level prediction, and the results of other works are also presented at the video-level.


\subsection{Intra-dataset Evaluation}

We conduct intra-dataset experiments on widely used FF++ dataset, including both high-quality (C23) and low-quality (C40) datasets. All models are trained and tested on the same dataset, where the performance reveals the model's capability in capturing the manipulation traces in forged faces. Table~\ref{intra-dataset} illustrates the intra-dataset results, where we bold and underline the best and the second-best scores, respectively.

In general, many manipulation traces are removed in C40 compressed data, making them hard to be detected. Under this challenging setting, it can be observed that our proposed FA-ViT outperforms most previous arts by a considerable margin. For example, our method surpass recent ViT-based model, F2Trans, by 2.53\% in terms of AUC. When evaluating on C23 data, all SOTA methods are able to achieve very high detection ACCs. FA-ViT still achieves outstanding 97.86\% ACC and 99.60\% AUC scores, demonstrating its effectiveness on C23 data. Moreover, compared with the ViT baseline, FA-ViT improves the ACC from 90.00\% to 92.17\% under the C40 setting and 96.00\% to 97.86\% under the C23 setting.

\begin{table}[]
 \centering
  \caption{Intra-dataset evaluation on FF++. '*' indicates the trained model provided by the authors. '\dag' indicates our re-implementation with the official code.}
  \label{intra-dataset}
\begin{tabular}{cccccc}
\toprule
\multirow{2}{*}{Method} & \multirow{2}{*}{Venue} & \multicolumn{2}{c}{FF++ (C40)} & \multicolumn{2}{c}{FF++ (C23)} \\ \cline{3-6} 
                        &                        & ACC          & AUC          & ACC          & AUC          \\ \hline
UIA-ViT \cite{zhuang2022uia}                    & ECCV 2022              & -        & -        & -        & 99.33        \\
F2Trans  \cite{miao2023f}              & TIFS 2023              & 90.19        & 94.11        &98.14       &\underline{99.73}        \\
DCL* \cite{sun2022dual}                     & AAAI 2022              & -            & -            & 97.62        & 99.57        \\
SBI* \cite{shiohara2022detecting}           & CVPR 2022              & -            & -            & 75.29        & 81.95        \\
LSDA* \cite{yan2024transcending}            & CVPR 2024              & -            & -            & 96.67        & 99.42        \\
CFM\dag \cite{luo2023beyond}                & TIFS 2023              & \textbf{94.57}        & \textbf{97.87}        & \textbf{98.54}        & 99.62        \\
RECCE\dag \cite{cao2022end}                 & CVPR 2022              & 91.59       & 96.05      & \underline{98.23}        & \textbf{99.76}        \\
ViT\dag \cite{dosovitskiy2021an}                 & ICLR 2021              & 90.00        & 93.49        & 96.00        & 98.92        \\ \midrule
FA-ViT                     & Ours                   &\underline{92.71}& \underline{96.64}        & 97.86        & 99.60        \\ \bottomrule
\end{tabular}
\end{table}

\subsection{Cross-dataset Evaluation}

Cross-dataset evaluation is a very challenging setting because the scenarios and characters in these datasets are complex and distinct to the training data. To comprehensively evaluate the generalization performance on different unseen datasets, we conduct extensive experiments on seven benchmark datasets. Specifically, all the models are trained on FF++ (C23) and evaluated on unforeseen datasets: CDF, WDF, DFDC-P, DFDC, DFD, DFR, and FFIW. 

Table~\ref{cross-dataset} illustrates the cross-dataset comparison results in terms of AUC. Compared to other ViT-based methods on CDF, including UIA-ViT, F2Trans, TALL-ViT and ViT, our proposed FA-ViT outperforms these methods by 4.98\%, 6.37\%, 7.25\%, and 10.05\%, respectively. Although CFM achieves the best intra-dataset performance, its generalization performance is inferior to that of our proposed FA-ViT. Moreover, FA-ViT achieves 4.88\% higher AUC than the recent method LSDA, highlighting the state-of-the-art generalization capability of FA-ViT for detecting unseen Deepfake faces. Overall, FA-ViT shows excellent performance across all seven datasets, these finds verifies the effectiveness of adaptive learning paradigm for pre-trained ViT.

\begin{table*}[]
  \caption{Cross-dataset evaluation on seven unseen datasets in terms of AUC. '*' indicates the trained model provided by the authors. '\ddag' indicates the results provided by the authors. '\dag' indicates our re-implementation.}
  \label{cross-dataset}
\centering
\scalebox{1.0}{
\begin{tabular}{cccccccccc}
\toprule
Method                          & Venue      & CDF   & WDF   & DFDC-P & DFDC  & DFD   & DFR   & FFIW  & Average \\ \hline
UIA-ViT\ddag \cite{zhuang2022uia}          & ECCV 2022  & 88.85 & - & 79.54      & -     & -     & -     & -     & -       \\
F2Trans \cite{miao2023f}     & TIFS 2023  & 87.46 & -     & 77.69  & -     & -     & -     & -     & -       \\
TALL-ViT \cite{xu2023tall} & ICCV 2023 & 86.58 & -     & -  & 74.10     & -     & -     & -     & -       \\
DCL* \cite{sun2022dual}         & AAAI 2022  & 88.24 & 77.57 & 76.87  & \underline{75.03} & 92.91 & 97.48 & 86.26 & 85.05        \\
SBI* \cite{shiohara2022detecting}          & CVPR 2022  & 88.61 & 70.27 & \underline{84.80}  & 71.70 & 82.68 & 89.85 & 86.62 & 82.07       \\
LSDA* \cite{yan2024transcending} & CVPR 2024 & 89.80 & 75.60     & 81.20  & 73.50     & \textbf{95.60}     & 89.20     & 87.90     & 84.69       \\
CFM\dag \cite{luo2023beyond}     & TIFS 2023  & \underline{89.65} & \underline{82.27} & 80.22  & 70.59 & \underline{95.21} & \underline{97.59} & 83.81 & \underline{85.62}         \\
RECCE\dag \cite{cao2022end}     & CVPR 2022  & 69.25 & 76.99 & 66.90  & 70.26 & 86.87 & 96.33 & \underline{91.20} & 79.68         \\
ViT\dag \cite{dosovitskiy2021an}   & ICLR 2021  & 83.78 & 78.04 & 78.53  & 73.95 & 91.73 & 94.64 & 76.92 & 82.51        \\ \hline
FA-ViT        & Ours       & \textbf{93.83} & \textbf{84.32} & \textbf{85.41}  & \textbf{78.32} & 94.88 & \textbf{98.01} & \textbf{92.22} &  \textbf{89.57}       \\ \bottomrule
\end{tabular}}
\end{table*}

\subsection{Cross-manipulation Evaluation}

The generalization performance of face forgery detectors on new manipulation methods is very important in real-world application. We follow the protocol proposed in \cite{sun2022dual, miao2023f} to evaluate cross-manipulation performance, where the model is trained on various manipulation types of samples and tested on unknown manipulation methods. The results are reported in Table~\ref{cross-mani}. It is observed that our proposed method achieves superior performance in cross-manipulation evaluation, exceeding the recent state-of-the-art method F2Trans by 3.22\% on GID-DF (C23) and 3.42\% on GID-DF (C40) in terms of ACC, respectively. Notably, we observe that the fully fine-tuned ViT does not generalize well to F2F, while our proposed FA-ViT achieves substantial improvements on both GID-F2F (C23) and GID-F2F (C40), demonstrating that adaptive learning strategy greatly enhance the model's generalization performance.


\begin{table*}[]
  \caption{Cross-manipulation evaluation on FF++, GID-DF means training on the other three manipulated methods while testing on deepfakes class. Results are cited from \cite{miao2023f}, '\dag' indicates our re-implementation.}
  \label{cross-mani}
\centering
\setlength{\tabcolsep}{8 pt}
\begin{tabular}{ccccccccc}
\toprule
\multirow{2}{*}{Method} & \multicolumn{2}{c}{GID-DF (C23)} & \multicolumn{2}{c}{GID-DF (C40)} & \multicolumn{2}{c}{GID-F2F (C23)} & \multicolumn{2}{c}{GID-F2F (C40)} \\ \cline{2-9} 
                        & ACC             & AUC            & ACC             & AUC            & ACC             & AUC             & ACC             & AUC             \\ \hline
EfficientNet \cite{tan2019efficientnet}            & 82.40           & 91.11          & 67.60           & 75.30          & 63.32           & 80.10           & 61.41           & 67.40           \\
MLGD \cite{li2018learning}                    & 84.21           & 91.82          & 67.15           & 73.12          & 63.46           & 77.10           & 58.12           & 61.70           \\
LTW \cite{sun2021domain}                     & 85.60           & 92.70          & 69.15           & 75.60          & 65.60           & 80.20           & 65.70           & 72.40           \\
DCL \cite{sun2022dual}                     & 87.70           & 94.90          & 75.90           & 83.82          & 68.40           & 82.93           & 67.85           & 75.07           \\
M2TR \cite{wang2022m2tr}                    & 81.07           & 94.91          & 74.29           & 84.85          & 55.71           & 76.99           & 66.43           & 71.70           \\
F3Net \cite{qian2020thinking}                  & 83.57           & 94.95          & 77.50           & 85.77          & 61.07           & 81.20           & 64.64           & 73.70           \\
F2Trans \cite{miao2023f}             & \underline{89.64}           & \underline{97.47}          & \underline{77.86}           & \underline{86.92}          & \underline{81.43}           & \underline{90.55}           & \underline{66.79}           & \underline{76.52}           \\ 
ViT\dag \cite{dosovitskiy2021an}      & 86.79           & 94.18          & 75.14           & 82.09          & 64.29           & 75.40           & 62.38           & 71.12           \\ \hline
FA-ViT                     & \textbf{92.86}           & \textbf{98.10}    & \textbf{81.28}     & \textbf{87.89}    & \textbf{82.57}     & \textbf{91.20}    &\textbf{67.91}     & \textbf{76.84}      \\ \bottomrule
\end{tabular}
\end{table*}

\begin{table}[]
 \centering
  \caption{Robust evaluation on real-world perturbations in terms of ACC. Results are cited from \cite{miao2023f}, '\dag' indicates our re-implementation.}
  \label{robustness}
\begin{tabular}{ccccc}
\toprule
\multirow{2}{*}{Method} & \multicolumn{4}{c}{std (training set)}    \\ \cline{2-5} 
                         & std/rand & std/sing & std/mix3 & std/mix4 \\ \hline
ResNet+LSTM \cite{jiang2020deeperforensics}             & 97.13    & 90.63    & -        & -        \\
F3Net \cite{qian2020thinking}                   & 92.79    & 86.82    & 79.10    & 66.42    \\
Xception \cite{chollet2017xception}                 & 94.75    & 88.38    & 82.32    & -        \\
M2TR \cite{wang2022m2tr}                     & 94.78    & 91.79    & 86.32    & 78.36    \\
MTD-Net \cite{yang2021mtd}  & 95.22    & -        & 86.89    & -        \\
F2Trans \cite{miao2023f}       &  \underline{97.26}    & \underline{95.77}    & \underline{93.28}    & \underline{90.55}    \\ 
ViT\dag \cite{dosovitskiy2021an}     & 96.77    & 93.53    & 90.05    & 87.06    \\ \midrule
FA-ViT                     & \textbf{98.76}    & \textbf{96.27}    & \textbf{94.03}    & \textbf{91.54}    \\ \bottomrule
\end{tabular}
\end{table}

\subsection{Robustness to Real-World Perturbations}

The transmission of Deepfake videos through online social networks inevitably introduces various distortions \cite{wu2022robust2}, such as video compression, noise, etc. To study the robustness performance on unseen perturbations, we follow the protocol proposed in DFR \cite{jiang2020deeperforensics}. Specifically, diverse perturbations are applied to the test set of DFR, including single-level random-type distortions (std/sing), random-level random-type distortions (std/rand), a mixture of three random-level random-type distortions (std/mix3), and a mixture of four random-level random-type distortions (std/mix4). We train the FA-ViT and the ViT on the pristine training set of DFR.

Table~\ref{robustness} illustrates the comparison results in terms of ACC. We can observe that other methods degrade dramatically when multiple types of perturbation are applied to pristine inputs in the std/mix3 and std/mix4 settings. For example, ViT achieves 96.77\% in the std/rand setting, but its performance drops to 87.06\% in the std/mix4 setting. In contrast, our proposed FA-ViT achieves the best performance across all distortion settings, with its performance declining more gradually in complex scenarios. The primary reason may be that we preserve the pre-trained ViT features in our adaptive learning process, as these features have demonstrated robust performance under various perturbations \cite{li2023trade}.

\subsection{Ablation Studies}
\label{section_ablation}
To analyze the impacts of different components in our proposed FA-ViT, we conduct ablation experiments by training all variants on FF++ (C23). We present the intra-dataset results on FF++ and cross-dataset results on CDF and WDF datasets. To ensure fair comparisons, all experiments are conducted using the same random seed.

\begin{figure}[]
  \centering
   \includegraphics[width=3.4 in]{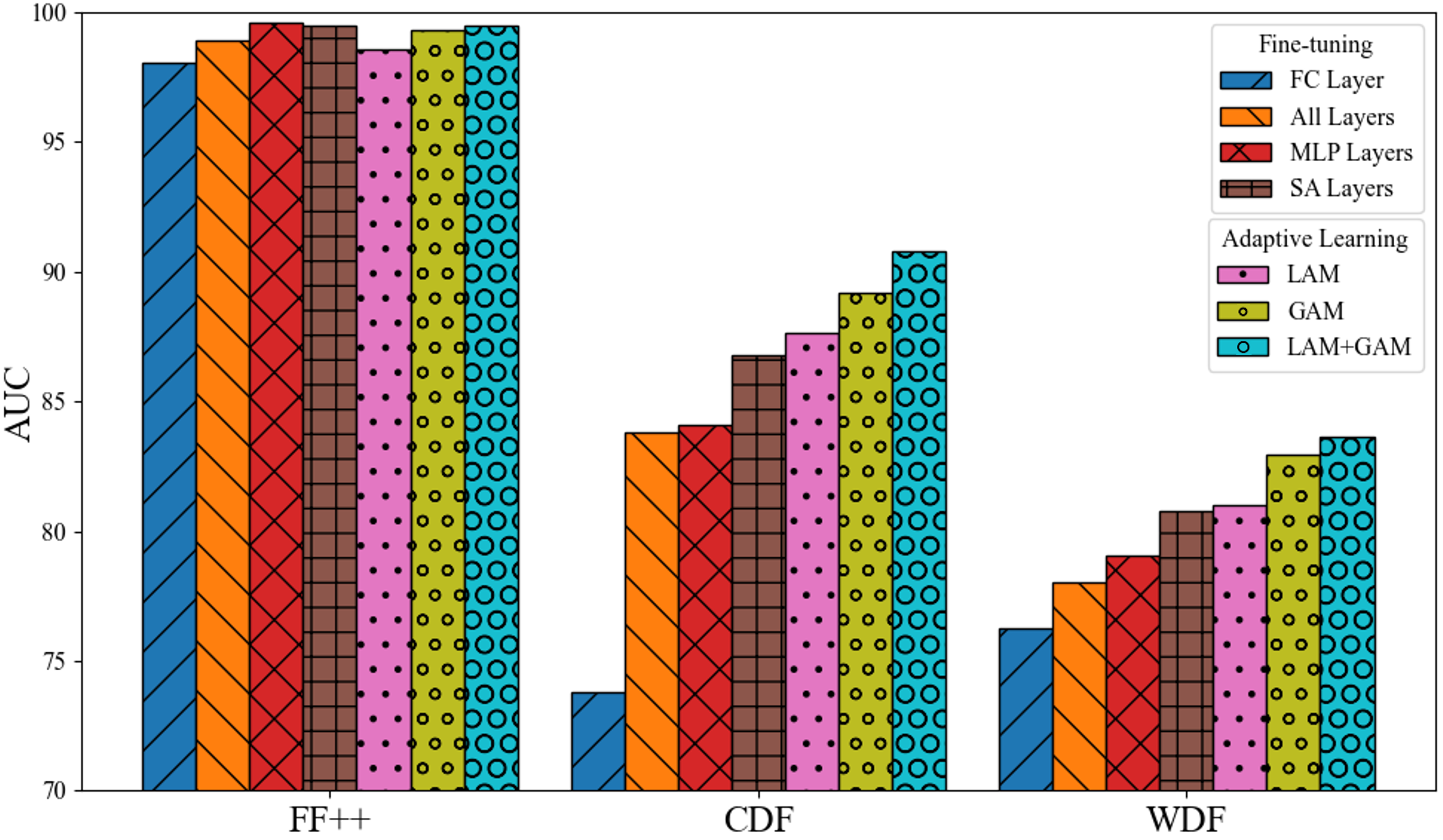}

   \caption{The comparison between different model adaptations in terms of AUC.}
   \label{model_adaptation}
\end{figure}

\begin{table}[]
\centering
\caption{Ablation on freezing pre-trained parameters.}
\label{freezing}
\begin{tabular}{ccccc}
\toprule
\multirow{2}{*}{Method}          & \multicolumn{2}{c}{FF++}      & CDF          &WDF      \\ \cline{2-5}
             & ACC         & AUC        & AUC       & AUC   \\ \hline 
w/o freezing      & 95.43         & 99.14           & 86.84    & 79.12    \\
w/ freezing        & \textbf{97.29} & \textbf{99.51} & \textbf{90.78}   &\textbf{83.63}      \\ \bottomrule
\end{tabular}
\end{table}

\subsubsection{Impacts of Different Model adaptations} 
\label{section_model_ablation}
To validate the superiority of the proposed adaptive learning paradigm, we apply various model adaptation strategies to the same pre-trained ViT with CE loss. The results shown in Fig~\ref{model_adaptation} lead to the following observations. (1) Fine-tuning only the last FC layer while keeping the pre-trained ViT fixed yields relatively high detection performance on FF++, but this strategy does not generalize well to CDF and WDF. This indicates that while pre-trained knowledge is feasible for detecting Deepfakes, it does not provide a generalized forensic representation. Therefore, it is necessary to learn additional forgery-aware information to enhance the model's expressiveness. (2) Partially fine-tuning SA layers or MLP layers achieves better generalization performance than the fully fine-tuning strategy. However, partially fine-tuning strategy still disrupt the pre-trained feature, resulting in inferior generalization performance compared to the proposed adaptive learning strategy. (3) Adaptive learning paradigm for pre-trained ViT achieves comparatively high performance, and the best performance is observed when both LAM and GAM collaborate to capture different types of forgery-aware information. Furthermore, as presented in Table ~\ref{freezing}, allowing ViT parameters to update during the adaptive training process leads to significant performance drops. This demonstrates that disrupting the pre-trained features results in sub-optimal generalization performance from another perspective.

\begin{table}[]
\centering
\caption{The comparison between different adaptations layer.}
\label{adaptation_layer}
\begin{tabular}{ccccc}
\toprule
\multirow{2}{*}{\begin{tabular}[c]{@{}c@{}}Adaptation\\ Layer\end{tabular}}         & \multicolumn{2}{c}{FF++}      & CDF          &WDF      \\ \cline{2-5}
                         & ACC         & AUC        & AUC       & AUC   \\ \hline
MLP                      & 96.14       & 99.07      & 88.35     & 81.78 \\
SA                    & \textbf{97.86}       & \textbf{99.60}      & \textbf{93.83}     & \textbf{84.32} \\ \bottomrule
\end{tabular}
\end{table}

\subsubsection{Ablation of GAM}
While the proposed GAM is designed to be inserted into the SA layer, another option is to insert it into the MLP layer. Table ~\ref{adaptation_layer} presents a comparison of inserting GAM into different layer of ViT. It can be observed that insert the proposed GAM into SA layer achieves better performance in both intra- and cross-dataset evaluation. This is because GAM is specifically designed to interact with query, key, and value tokens, allowing it to leverage the self-attention mechanism to effectively capture global forgery-aware information.

\begin{table}[]
\centering
\caption{Comparison of different adaptation method for spatial information in FA-ViT. '-' means without adapting spatial information to ViT.}
\label{componets ablation}
\begin{tabular}{ccccc}
\toprule
\multirow{2}{*}{Method}          & \multicolumn{2}{c}{FF++}      & CDF          &WDF      \\ \cline{2-5}
                         & ACC         & AUC        & AUC       & AUC   \\ \hline
-                     & 97.43       & 99.45      & 90.33     & 82.60 \\   
Add                      & 96.98       & 99.39      & 91.41     & 80.57 \\
CA                       & 97.57       & 99.55      & 92.31     & 82.15 \\
LAM                     & \textbf{97.86}       & \textbf{99.60}      & \textbf{93.83}     & \textbf{84.32} \\ \bottomrule
\end{tabular}
\end{table}

\begin{figure}[]
  \centering
   \includegraphics[width=3.4 in]{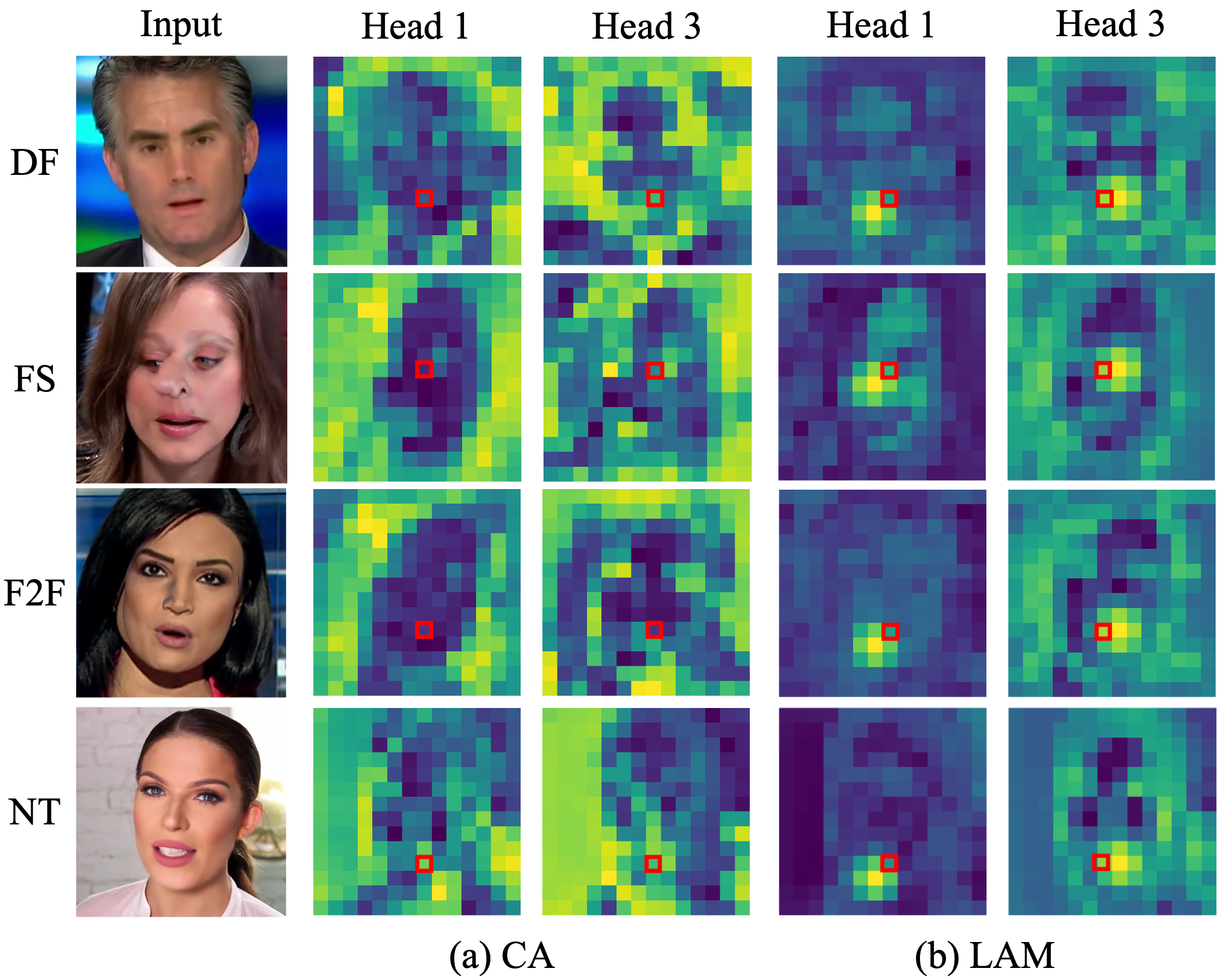}

   \caption{Attention maps from (a) Cross Attention (CA) and (b) LAM, the query position is marked by the red box.}
   \label{att_map}
\end{figure}

\subsubsection{Effectiveness of LAM} 
To examine the effectiveness of LAM, we conduct a comparison with two commonly used methods for incorporating spatial information into ViT: (1) directly adding spatial information to each ViT token \cite{miao2023f}; and (2) using the Cross Attention (CA) module to attentively inject spatial information into ViT \cite{song2022face}. As demonstrated in Table.~\ref{componets ablation}, using the addition operation does not bring consistent improvements across different datasets, while CA module is not effective on WDF dataset. In contrast, LAM achieves the improvements of 3.50\% on CDF and 1.72\% on WDF, respectively. Unlike previous works \cite{song2022face, miao2023f}, where ViT parameters are allowed to collaboratively update to optimize the fusion process, adaptive learning requires the designed module to establish an effective connection between each query token and the spatial information. LAM considers the local context prior in spatial space and optimizes the combination of global and local relations in this process, thereby effectively adapting local spatial information to pre-trained ViT. Furthermore, we visualize the attention map from CA and LAM, as shown in Fig.~\ref{att_map}. This visualization clearly shows that LAM specifically highlights the neighboring relations surrounding the query token and simultaneously interacts with long-range spatial information, demonstrating its effectiveness from another point of view.

\begin{figure*}[]
  \centering
  \includegraphics[scale=0.4]{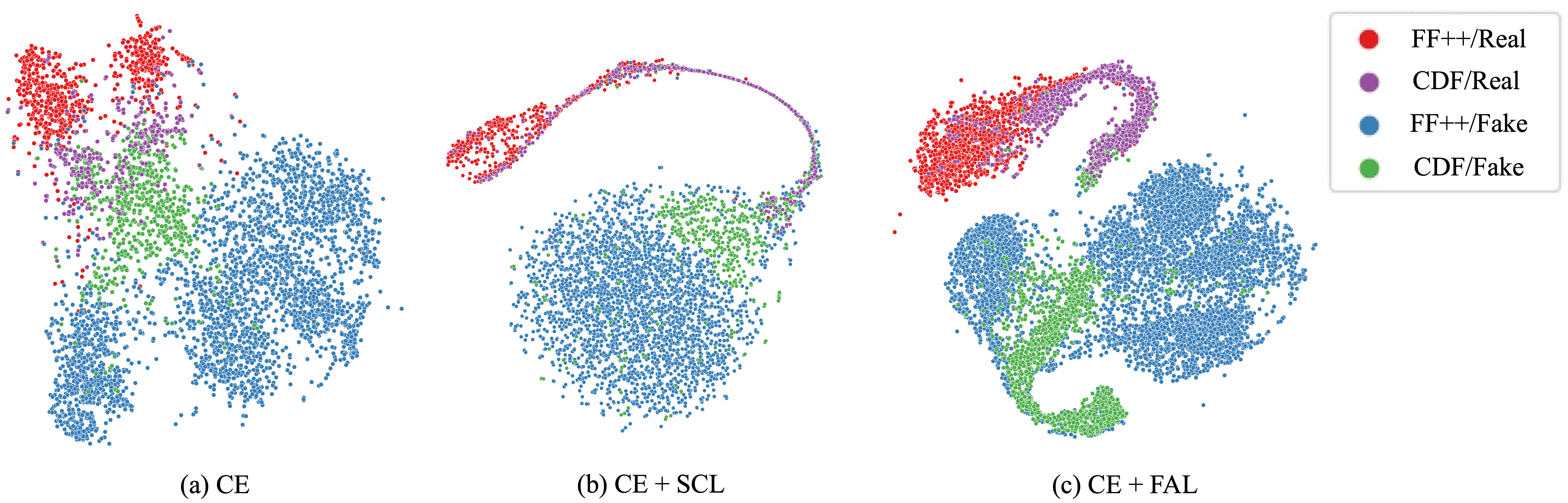}

   \caption{t-SNE feature distribution of our model by supervised with (a) Cross Entropy (CE) loss, (b) CE loss and Single Center Loss (SCL) \cite{li2021frequency}, (c) CE loss and our proposed FAL.}
   \label{tsne}
\end{figure*}

\begin{table}[]
\centering
\caption{Ablation experiment on the proposed FAL. Variation 1 uses triplet loss to replace circle loss in FAL, while variant 2 removes the constraint that applies only to fine-grained pairs }
\label{FAL ablation}
\begin{tabular}{ccccc}
\toprule
\multirow{2}{*}{Varation}          & \multicolumn{2}{c}{FF++}      & CDF          &WDF      \\ \cline{2-5}
             & ACC         & AUC        & AUC       & AUC   \\ \hline 
Variation 1      & 97.58         & 99.52           & 92.54    & 83.56    \\             
Variation 2     & \textbf{98.14}         & \textbf{99.82}           & 91.28    & 81.71    \\
FAL        & 97.86 & 99.60 & \textbf{93.83}   &\textbf{84.32}      \\ \bottomrule
\vspace{-5 mm}
\end{tabular}
\end{table}

\begin{table}[]
\centering
\caption{Comparison experiment on the proposed FAL.}
\label{loss camparasion}
\begin{tabular}{ccccc}
\toprule
\multirow{2}{*}{\begin{tabular}[c]{@{}c@{}}Loss\\ Function\end{tabular}} & \multicolumn{2}{c}{FF++}      & CDF          &WDF      \\ \cline{2-5}
            & ACC         & AUC        & AUC       & AUC   \\ \hline 
CE       & 97.29     & 99.51        & 90.78          & 83.63 \\      
CE + SCL \cite{li2021frequency}      & 97.43     & 99.46        & 91.67          & 80.60 \\
CE + FAL   & \textbf{97.86}     & \textbf{99.60}        & \textbf{93.83}          & \textbf{84.32} \\ \bottomrule
\end{tabular}
\end{table}

\subsubsection{Experiments on the Proposed FAL} 

As described in Sec.~\ref{intro_FAL}, FAL is designed to expose subtle forgery discrepancies in each fine-grained pair and provides flexible optimization for model updates. We conduct the ablation experiment on FAL by designing two variations: (1) Variation 1 replaces circle loss with triplet loss; and (2) Variation 2 extends FAL to include non-fine-grained pairs. As the results illustrated in Table ~\ref{FAL ablation}, Variation 1 shows inferior performance compared to the proposed FAL. This is because circle loss performs adaptive weighting for each fine-grained pair, thereby enabling the model to learn a better representation. Meanwhile, it can be observed that Variation 2 achieves slight improvements on intra-dataset evaluation but shows a heavy degradation on unseen datasets. Considering the significant differences in non-fine-grained pairs, the overfitting problem may arise from learning trivial features in non-essential regions.

Additionally, we compare the proposed FAL with SCL \cite{li2021frequency}, which regularizes the mean distance between different categories to compact the intra-class variant. As illustrated in Table.~\ref{loss camparasion}, although SCL shows improvements on CDF, its performance on WDF decreases. In contrast, we can observe FAL achieves an improvement on CDF from 90.78\% to 93.83\%, and on WDF from 83.63\% to 84.32\%. Furthermore, the t-SNE feature distributions \cite{van2008visualizing} under different optimization objects are shown in Fig.~\ref{tsne}. Compared to SCL, the proposed FAL exhibits a more compact intra-class distribution and a more discriminative inter-class distribution, thus demonstrating its superiority in the case of adaptive learning paradigm.

\begin{table}[]
\caption{Impacts of different combination of $m$ and $\eta$ in terms of AUC.}
\label{parameter}
\centering
\begin{tabular}{ccccc}
\toprule
\multicolumn{2}{c}{Hyper-parameters} & \multirow{2}{*}{FF++} & \multirow{2}{*}{CDF} & \multirow{2}{*}{WDF} \\ \cline{1-2}
$m$  & $\eta$  &                       &                      &                      \\ \hline
0       &24            &\textbf{99.82}                 &90.40                 &81.56                      \\
0.25    &24            &99.60         &\textbf{93.83}       &\textbf{84.32}         \\ 
0.5     &24            &99.59                  &92.90                 &82.43                     \\ \hline
0.25    &1             &99.44                  &92.10                 &82.48                      \\ 
0.25    &24            &99.60                  &\textbf{93.83}                 &\textbf{84.32}                      \\
0.25    &32            &\textbf{99.61}                  &92.59                 &83.37                      \\ \bottomrule
\end{tabular}
\end{table}

\subsubsection{Impacts of Different Parameters in FAL}
In FAL, the parameter $m$ controls the decision boundary's radius, while $\eta$ serves as a scaling factor. Table~\ref{parameter} presents the intra- and cross-dataset performance under various combinations of $m$ and $\eta$. Initially, we set $\eta$ to 24 and examine the impacts of different $m$. We observe that setting $m$ to 0 improves intra-dataset performance but hinders its generalization. This highlights the risk of overfitting when all genuine faces are excessively concentrated towards the center point. Consequently, we find that introducing a radius by setting $m=0.25$ yields optimal performance. We then keep $m$ at 0.25 and investigate the impact of $\eta$. We observe that $\eta=1$ results in the sub-optimal performance, whereas increasing it to $\eta=24$ produces the best results. Finally, the hyper-parameters of $m$ and $\eta$ in FAL are set as 0.25 and 24, respectively.


\begin{table}[]
\centering
\caption{Impacts of different pre-trained initialization.}
\label{pre-trained}
\begin{tabular}{cccccc}
\toprule
\multirow{2}{*}{\begin{tabular}[c]{@{}c@{}}Pre-trained\\ Initialization\end{tabular}} & \multirow{2}{*}{FA-ViT} & \multicolumn{2}{c}{FF++}      & CDF          &WDF      \\ \cline{3-6}
          &               & ACC         & AUC        & AUC       & AUC   \\ \hline 
DeiT & -     & \textbf{96.00}     & \textbf{98.67}        & 82.99          &80.45  \\
\cellcolor[HTML]{E0DBDB}DeiT & \cellcolor[HTML]{E0DBDB}\checkmark & \cellcolor[HTML]{E0DBDB}95.71& \cellcolor[HTML]{E0DBDB}98.58 & \cellcolor[HTML]{E0DBDB}\textbf{87.50}  & \cellcolor[HTML]{E0DBDB}\textbf{84.82}\\ \hline
CLIP & -     & 95.86     & 98.84        & 82.61          &80.51  \\
\cellcolor[HTML]{E0DBDB}CLIP & \cellcolor[HTML]{E0DBDB}\checkmark & \cellcolor[HTML]{E0DBDB}\textbf{97.13}& \cellcolor[HTML]{E0DBDB}\textbf{99.33} & \cellcolor[HTML]{E0DBDB}\textbf{90.41}  & \cellcolor[HTML]{E0DBDB}\textbf{80.81}\\ \hline
ImageNet-1k & -      &95.86      & 98.85        & 82.76          &79.92  \\
\cellcolor[HTML]{E0DBDB}ImageNet-1k & \cellcolor[HTML]{E0DBDB}\checkmark & \cellcolor[HTML]{E0DBDB}\textbf{96.34}& \cellcolor[HTML]{E0DBDB}\textbf{99.48} & \cellcolor[HTML]{E0DBDB}\textbf{92.14} & \cellcolor[HTML]{E0DBDB}\textbf{80.85}\\ \hline
ImageNet-21K & -      &96.00      & 98.92        & 83.78          & 78.04  \\
\cellcolor[HTML]{E0DBDB}ImageNet-21K & \cellcolor[HTML]{E0DBDB}\checkmark & \cellcolor[HTML]{E0DBDB}\textbf{97.86}& \cellcolor[HTML]{E0DBDB}\textbf{99.60} & \cellcolor[HTML]{E0DBDB}\textbf{93.83} & \cellcolor[HTML]{E0DBDB}\textbf{84.32}\\ \bottomrule
\end{tabular}
\end{table}

\subsubsection{Impact of Different Pre-trained Initialization} We initialize ViT with different common used pre-trained weights, including DeiT \cite{touvron2021training}, CLIP \cite{radford2021learning}, ImageNet-1K \cite{deng2009imagenet}, and ImageNet-21K \cite{ridnik2021imagenet}. The results are illustrated in Table ~\ref{pre-trained}. It is evident that FA-ViT improve the generalization performance on different pre-trained ViTs. On the other hand, we note that different initialization significantly impacts generalization performance, indicating the importance of choosing a suitable pre-trained initialization. Empirically, we find that ViT trained on ImageNet-21K shows the highest performance for face forgery detection.

\subsubsection{Impacts of Different ViT Backbones} Herein, we study the impacts of different ViT backbones, including ViT-Small, ViT-Base, and ViT-Large. The results are illustrated in Table ~\ref{backbone}. Compared with directly finetuning the vanilla ViT backbones, our proposed FA-ViT consistently achieves significant improvements on generalization performance across these ViT backbones. Specially,  when we apply adaptive learning to the ViT-Large backbone, the model achieves 9.49\% and 2.27\% AUC boosts on CDF and WDF. According to this ablation experiment, we finally select ViT-Base as our basic backbone.

\begin{table}[]
\centering
\caption{Impacts of different ViT backbones.}
\label{backbone}
\begin{tabular}{cccccc}
\toprule
\multirow{2}{*}{Backbone} & \multirow{2}{*}{FA-ViT} & \multicolumn{2}{c}{FF++}      & CDF          &WDF      \\ \cline{3-6}
          &               & ACC         & AUC        & AUC       & AUC   \\ \hline 
ViT-Small & -     & 94.92     & 98.56        & 81.33          & 77.21 \\
\cellcolor[HTML]{E0DBDB}ViT-Small & \cellcolor[HTML]{E0DBDB}\checkmark & \cellcolor[HTML]{E0DBDB}\textbf{96.57}& \cellcolor[HTML]{E0DBDB}\textbf{99.41} & \cellcolor[HTML]{E0DBDB}\textbf{90.06}  & \cellcolor[HTML]{E0DBDB}\textbf{82.27}\\ \hline
ViT-Base & -      &96.00      & 98.92        & 83.78          & 78.04  \\
\cellcolor[HTML]{E0DBDB}ViT-Base & \cellcolor[HTML]{E0DBDB}\checkmark & \cellcolor[HTML]{E0DBDB}\textbf{97.86}& \cellcolor[HTML]{E0DBDB}\textbf{99.60} & \cellcolor[HTML]{E0DBDB}\textbf{93.83} & \cellcolor[HTML]{E0DBDB}\textbf{84.32}\\ \hline
ViT-Large & -      &95.43      & 98.93        & 83.44          &79.74  \\
\cellcolor[HTML]{E0DBDB}ViT-Large & \cellcolor[HTML]{E0DBDB}\checkmark & \cellcolor[HTML]{E0DBDB}\textbf{97.47}& \cellcolor[HTML]{E0DBDB}\textbf{99.61} & \cellcolor[HTML]{E0DBDB}\textbf{92.93} & \cellcolor[HTML]{E0DBDB}\textbf{82.01}\\ \bottomrule
\end{tabular}
\vspace{-3mm}
\end{table}

\begin{figure}[]
  \centering
   \includegraphics[width=3.2 in]{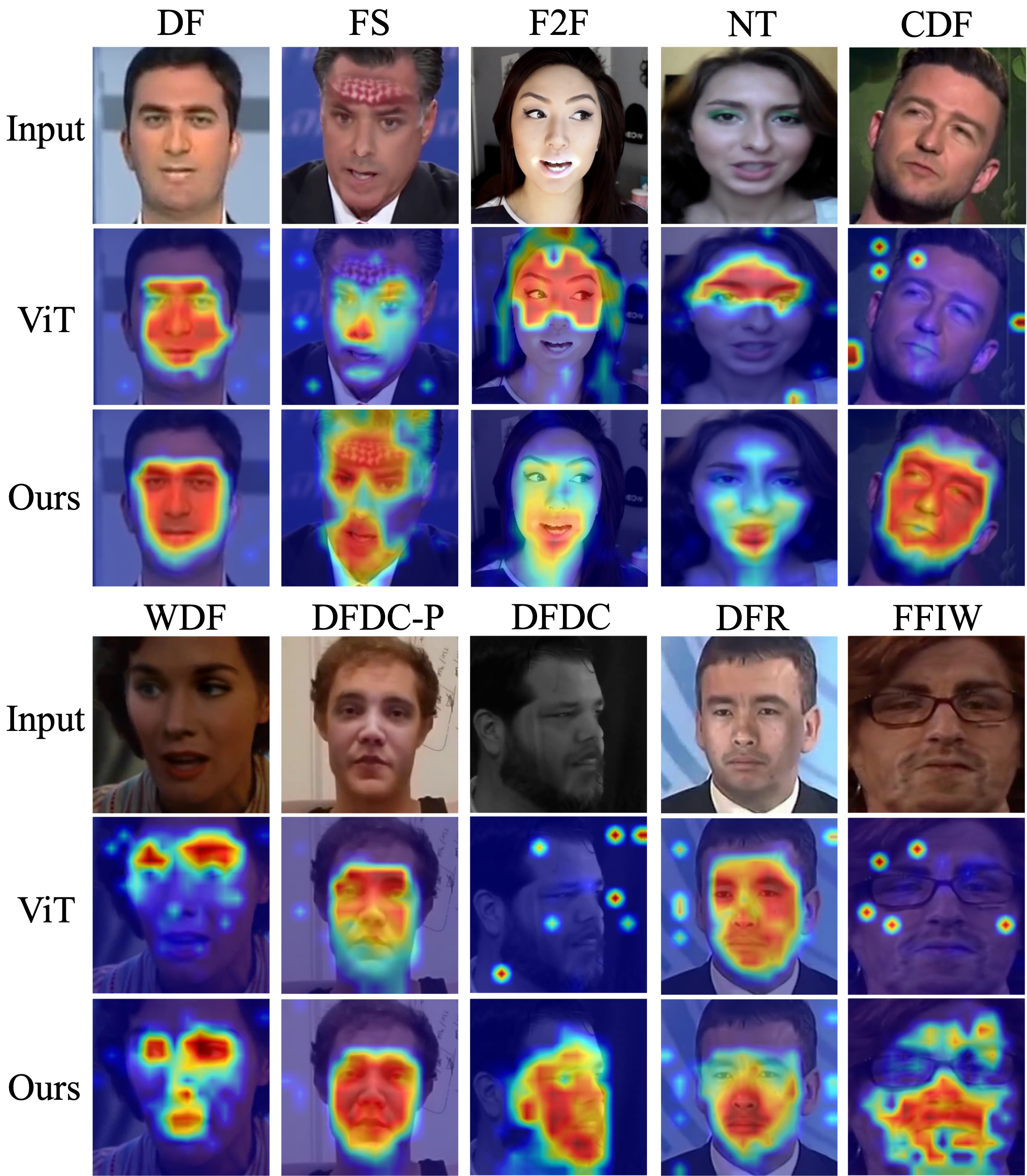}

   \caption{Saliency map visualization of ViT and our proposed FA-ViT on different dataset. }
   \label{Grad_cam}
    \vspace{-5mm}
\end{figure}

\subsection{Saliency Map Visualization}

To better clarify the effectiveness of our FA-ViT, we visualize the model's attention on Deepfake faces using Grad-CAM++ \cite{chattopadhay2018grad}, as shown in Fig.~\ref{Grad_cam}. It can be observed that FA-ViT generates distinguishable saliency maps for different Deepfake faces, and captures method-specific artifacts, such as the forehead region in FS and mouth region in F2F. In cross-dataset scenarios, ViT struggles to detect Deepfakes in complex environment, such as large-pose face in CDF or challenging lighting condition in DFDC. In contrast, FA-ViT consistently traces the manipulated regions across different unseen datasets, thus verifying its effectiveness from the decision-making perspective.

\section{Conclusion}

In this paper, we present a novel Forgery-Aware Adaptive Vision Transformer (FA-ViT) tailored for generalized face forgery detection from the perspective of model adaptation, which is somewhat ignored in previous study. Specifically, the expressivity of the pre-trained ViT is preserved during training, while we design Global Adaptive Module (GAM) and Local Adaptive Module (LAM) to adapt global and local forgery-aware information for generalized representation learning. In addition, we also introduce Fine-grained Adaptive Learning (FAL) to facilitate the adaptive learning of fine-grained forgery-aware information. In conclusion, our proposed framework offers a generalized and robust solution to the challenges of face forgery detection. We believe our proposed method can provide valuable insights to the research community, and further advance the development of face forgery detection systems.

\ifCLASSOPTIONcaptionsoff
  \newpage
\fi

\bibliographystyle{IEEEtran}
\bibliography{refers}

\end{document}